\documentclass[11pt]{article} 
\usepackage{rldmsubmit,palatino}
\usepackage{graphicx}
\usepackage{amsmath,amssymb,amsfonts}
\DeclareMathOperator*{\argmax}{arg\,max}

\usepackage{algorithmic}
\usepackage{graphicx}
\usepackage{textcomp}
\usepackage{url}

\title{Focusing Robot Open-Ended Reinforcement Learning\\ Through Users' Purposes}

\author{
Emilio Cartoni\thanks{
Emilio Cartoni: \url{https://www.istc.cnr.it/it/people/emilio-cartoni}. 
Gianluca Cioccolini: \url{https://www.istc.cnr.it/it/user/12865}. 
Gianluca Baldassarre: \url{https://www.istc.cnr.it/it/people/gianluca-baldassarre}.}, Gianluca Cioccolini, Gianluca Baldassarre \\
Laboratory of Embodied Natural and Artificial Intelligence (LENAI),\\
Institute of Cognitive Sciences and Technologies (ISTC),\\
National Research Council (CNR),\\
Rome, Italy\\
\texttt{\{emilio.cartoni, gianluca.cioccolini, gianluca.baldassarre\}@istc.cnr.it}\\
}

\begin{document}

\maketitle

\begin{abstract}

Open-Ended Learning (OEL) autonomous robots can acquire new skills and knowledge through direct interaction with their environment, relying on mechanisms such as intrinsic motivations and self-generated goals to guide learning processes.
OEL robots are highly relevant for applications as they can autonomously leverage acquired knowledge to perform tasks beneficial to human users in unstructured environments, addressing challenges unforeseen at design time.
However, OEL robots face a significant limitation: their openness may lead them to waste time learning information that is irrelevant to tasks desired by specific users.
Here, we propose a solution called `Purpose-Directed Open-Ended Learning' (POEL), based on the novel concept of `purpose' introduced in previous work.
A purpose specifies what users want the robot to achieve.
The key insight of this work is that purpose can focus OEL on learning self-generated classes of tasks that, while unknown during autonomous learning (as typical in OEL), involve objects relevant to the purpose.
This concept is operationalised in a novel robot architecture capable of receiving a human purpose through speech-to-text, analysing the scene to identify objects, and using a Large Language Model to reason about which objects are purpose-relevant.
These objects are then used to bias OEL exploration towards their spatial proximity and to self-generate rewards that favour interactions with them.
The solution is tested in a simulated scenario where a camera-arm-gripper robot interacts freely with purpose-related and distractor objects.
For the first time, the results demonstrate the potential advantages of purpose-focused OEL over state-of-the-art OEL methods, enabling robots to handle unstructured environments while steering their learning toward knowledge acquisition relevant to users.

\end{abstract}


\acknowledgements{This work has received funding from: the European Union’s Horizon 2020 Research and Innovation Programme, GA No 101070381, project 'PILLAR-Robots - Purposeful Intrinsically motivated Lifelong Learning Autonomous Robots';
the ‘European Union, NextGenerationEU, PNRR’, project ‘EBRAINS-Italy - European Brain ReseArch INfrastructureS Italy’, MUR code IR0000011, CUP B51E22000150006 and project ‘FAIR - Future Artificial Intelligence Research’, MUR code PE0000013, CUP B53C22003630006.}

\startmain 

\section{Introduction}
This work proposes a specific solution that instantiates the concept of \textit{purpose}, a novel means to exploit the capacity of open-ended learning (OEL) robots to autonomously seek and learn to solve tasks while at the same time orienting their self-driven knowledge acquisition towards tasks relevant to given users.

\noindent
\textbf{OEL: objectives and potential.} Autonomous OEL robots can cumulatively acquire new skills and knowledge through direct interaction with the environment, for example by relying on the guidance of \textit{intrinsic motivations} (e.g., novelty, surprise, and the acquisition of competence) and the \textit{self-generation of goals} guiding skill acquisition \cite{SantucciBaldassarreMirolli2013Whichisthebestintrinsicmotivationsignalforlearningmultipleskills,DoncieuxFilliatDiazRodriguezHospedalesDuroConinxRoijersGirardPerrinSigaud2018OpenEndedLearningaConceptualFrameworkBasedonRepresentationalRedescription,SigaudBaldassarreColasDoncieuxDuroPerrinGilbertSantucci2023ADefinitionOfOpenEndedLearningProblemsForGoalConditionedAgents}. 
These mechanisms allow robots to acquire knowledge and skills before knowing the tasks they will have to address in the future \cite{DoncieuxFilliatDiazRodriguezHospedalesDuroConinxRoijersGirardPerrinSigaud2018OpenEndedLearningaConceptualFrameworkBasedonRepresentationalRedescription,Baldassarre2013Book,CartoniMontellaTrieschBaldassarre2020AnOpenEndedLearningArchitecturetoFacetheREAL2020SimulatedRobotCompetition}.
In this respect, we have previously proposed the following objective function for OEL.
The key idea behind it is to measure the quality of knowledge acquired through autonomous \textit{intrinsic exploration} based on how it effectively supports the later solution of \textit{extrinsic} tasks \cite{CartoniMontellaTrieschBaldassarre2020AnOpenEndedLearningArchitecturetoFacetheREAL2020SimulatedRobotCompetition}:
\begin{equation}
  \theta^* = \argmax_\theta E_{g\sim\tau(g)} \left(E_{\pi(a|s,g,\theta)}R(g)\right) 
\end{equation}
where
$\theta$ are the parameters of the robot controller to optimise,
$g \sim \tau(g)$ are a sample of goals drawn from the distribution of all possible goals in the given environment,
$R(g)$ is the total reward measuring the performance of goal $g$,
$\pi(a|s,g,\theta)$ is the robot goal-conditioned action policy selecting action $a$ in response to the state $s$ and goal $g$. 
The robot searches for the optimal parameters $\theta^*$ during an initial autonomous exploration phase, should ideally allow it to later solve \textit{any} possible goal in the given environment.
OEL robots have great potential for applications as they can use autonomously acquired skills and knowledge to perform tasks relevant to human users \cite{CartoniMontellaTrieschBaldassarre2020AnOpenEndedLearningArchitecturetoFacetheREAL2020SimulatedRobotCompetition,Seepanomwan2017}.
The autonomy of OEL robots 
allows their employment to solve tasks in unstructured environments posing challenges that are impossible to foresee at design time.
For example, these are environments where humans live, for example houses, offices, and workshops.

\noindent
\textbf{OEL: limitations addressed here.}
An important limitation of OEL robots is that they get hooked on any possible experience deemed interesting. 
For example, the intrinsic motivations of robots might tend to drive them to focus on learning any goal that increases their competence \cite{Santucci2016,NairPongDalalBahlLinLevine2018VisualReinforcementLearningwithImaginedGoals}.
As a consequence, given that the time available for free autonomous exploration is limited, OEL robots tend to acquire shallow knowledge and skills over \textit{all} possible tasks they discover, including those that are irrelevant to their users.

\noindent
\textbf{Proposed solution: Purpose-directed Open Ended Learning (POEL).}
This work proposes a possible solution to the OEL unfocused exploration problem by pivoting on the novel concept of \textit{purpose}, on which we have recently proposed a computational framework  \cite{BaldassarreDuroCartoniKhamassiRomeroSantucci2024PurposeForOpenEndedLearningRobotsAComputationalTaxonomyDefinitionAndOperationalisation}.
A purpose indicates what the designer and/or user want from the robot.
The robot should receive or acquire the purpose, have a suitable representation of it, and use it to direct its autonomous learning processes towards the solution of specific tasks that are relevant for the purpose.
While our previous work developed a theoretical and formal framework on purpose, here we propose possible ways to use purpose to bias autonomous OEL.
In particular, among the possible uses of purpose \cite{BaldassarreDuroCartoniKhamassiRomeroSantucci2024PurposeForOpenEndedLearningRobotsAComputationalTaxonomyDefinitionAndOperationalisation}, we consider here the case where purpose indicates a general \textit{class of possible tasks} from which the user might later select specific ones.
In particular, we propose a solution is grounded on the insight for which \textit{different classes of purpose-relevant tasks involve the use of different classes of objects}.
Based on this, we propose solutions where the robots' OEL exploration processes are biased towards \textit{purpose-relevant objects}.
For example, a user might want the robot to carry out the class of `tasks with toys', for example, gathering toys from the floor, sorting toys into containers, putting toys on a shelf in a certain order, etc.
Another user might want the robot to carry out a different class of `tasks with fruit', for example, collecting fruit from trees, packaging fruit, piling fruit on a table, etc. 
The general idea is that in these cases the robot should focus its OEL exploration towards the acquisition of knowledge towards objects that are relevant for the purpose.
Within these boundaries, however, the robot should still be able to leverage OEL autonomy to acquire knowledge that is potentially relevant to solve all future user's tasks related to the purpose.
This work specifically proposes a novel \textit{Purpose-directed Open Ended Learning} (POEL) robot architecture that receives a human purpose through speech-to-text, analyses the scene to identify objects, and uses a Large Language Model (LLM) to reason about which objects are relevant for the purpose.
The objects' position is then used to bias OEL exploration in their spatial proximity and to self-generate rewards favouring interactions with them.

\noindent
\textbf{Contributions.}
Overall, the work makes the following contributions:
\begin{itemize}
  \item 
    It proposes a specific novel solution to focus the OEL of robots based on the novel concept of purpose; in particular it uses a LLM to identify purpose-relevant objects in the scene and to bias OEL onto them.
  \item 
    It tests the solution in a novel testbed scenario involving purpose-relevant and distractor objects.
  \item 
    It shows how POEL outperforms state-of-the-art OEL models which explore the environment in unfocused ways.
\end{itemize}


\section{Methods}
\label{Sec:Methods}

\paragraph{Building on LEXA RL model.}
POEL builds on LEXA, a state-of-the-art OEL model \cite{mendonca2021discovering}.
LEXA comprises key components, each with a distinct role.
The \textit{Replay Buffer} acts as a memory buffer that stores the agent's experiences used for training.
The \textit{World Model} learns the dynamics of the environment, in particular to predict the outcome of actions, and it is used to simulate future states of the environment in planning processes.
The \textit{Policy Explorer} guides the exploration of the environment to maximize intrinsic reward. 
This is a measure of the novelty of experiences estimated with the disagreement of an ensemble of experts that predict the consequences of actions given a world state.
A \textit{Policy Achiever} is a goal-conditioned policy that learns to achieve target goals. 
Goals are provided as images from the \textit{Replay Buffer}, and the policy is trained to reach the states represented in such images.
This allows the agent to learn how to accomplish various tasks by progressively exploring the environment and learning to achieve different goal states.

\paragraph{ALAN. } ALAN \cite{MendoncaBahlPathak2023ALANAutonomouslyExploringRoboticAgentsInTheRealWorld}, a successor of LEXA, is capable of efficiently learning real-world tasks on a real robot.
Unlike LEXA, its exploration aims to maximise environmental change during exploration (excluding the robot motion) and it uses `visual priors' (an object detection algorithm) to initialise the robot's gripper near objects.

\paragraph{Purpose Integration.}
POEL integrates the concept of Purpose into LEXA.
This integration involves several additional components and processes.
A \textit{Speech-to-Text} component (SpeechRecognition \cite{zhang2017speechrecognition})
is used to allow a user to communicate a purpose to the robot, for example `Learn to manipulate blue objects'.
A \textit{Large Language Model} (LLM) (LLaVA-v1.5-7b-q4 \cite{liu2023improvedllava}) interprets the user's Purpose, provided in natural language, translating the user's instructions into actionable objectives for the robotic agent. A \textit{Visual Module} uses a YOLO v9 \cite{wang2024yolov9}, an object detection model, to interpret the scene. 
Trained within the simulation environment, the YOLO v9 model identifies objects and their locations, providing a detailed visual understanding of the surroundings. 
Two additional purpose-oriented components have been incorporated into the World Model: the \textit{Proximity Purpose Model} and \textit{Lifting Purpose Model}. 
The Proximity Purpose Model is trained on observations of the robot’s interactions with the environment. 
It evaluates whether the distance between the end effector and the target object is less than 15 cm. The Lifting Purpose Model assesses whether the target object has been raised by at least 5 cm along the z-axis compared to its initial position.
These purpose models help direct the robot's exploration toward skills that facilitate achieving the Purpose.
The Policy Explorer aims to maximise a reward composed of three equally weighted components: the reward from ensemble disagreement in the World Model (as in LEXA), the Proximity Purpose Model reward, and the Lifting Purpose Model reward.
These latter two rewards are binary rewards which are set to 1 only when their respective model says that either the object proximity or lifting has been achieved. 
The robot processes the user's Purpose through the following steps:
\begin{enumerate}
    \item \textbf{User Input:} the user communicates the Purpose to the robot through speech, and this is converted to text via Speech-to-Text.
    \item \textbf{Visual Interpretation:} the visual module processes the simulated environment to identify objects and their locations.
    \item \textbf{LLM Processing:} the LLM receives the user's Purpose and the visual description of the scenario from the visual module, i.e. the list of the detected objects.
    The LLM interprets the Purpose and determines which objects are relevant to achieving it, giving as output a subset of the objects detected in the scene.
    This list of objects biases the robot’s  interactions to favour these objects, guiding the agent to focus on tasks relevant for the Purpose. 
    \item \textbf{Initialization:} based on the LLM's interpretation, the robotic arm's initial position is set near a relevant object. This strategic initialization increases the likelihood of the agent interacting with objects pertinent to the user's Purpose, enhancing learning efficiency and later task performance.
    \item \textbf{Purpose Proximity and Lifting rewards:} using the visual module’s ability to determine the position of the target objects, both observations coming from the environment and from imagined rollouts internal to LEXA are analysed by the Proximity and Lifting Purpose models and are assigned rewards.
    This enhances the interactions with Purpose-related objects and the learning of Purpose-relevant tasks.

\end{enumerate}
By incorporating these advanced components and processes, the robotic agent can align its actions with the user's Purpose, thus significantly improving the effectiveness of its interactions within the environment. 
This optimisation accelerates skill acquisition, allowing the robot to perform specific user's goals related to the Purpose more efficiently.

\section{Results}
\label{Sec:Results}

\subsection{Experimental setup}
We evaluated the architecture in a simulated environment involving a \textit{Kuka LBR iiwa R800} robotic arm positioned in the centre of a table.
In this setup, the robot is surrounded by three cubes (red, green, and blue) and two boxes. The robot is controlled through joint position deltas and operates with five degrees of freedom (three were left fixed). 
The observations provided to the agent consist of RGB images with dimensions 64×64×3.

The robot performance was evaluated with 24 distinct goals, each defined as an image, grouped as follows:
6 Posture Goals -- assuming specific postures with the arm;
2 Reaching Goals -- reaching the blue cube or the green cube;
8 Pushing Goals -- pushing the green or the blue cube;
8 Pick-and-Place Goals -- picking and placing the green or the blue cube. 

%
\begin{figure}[h!]
    \centering
    \begin{minipage}{0.44\textwidth}
        \includegraphics[width=\textwidth]{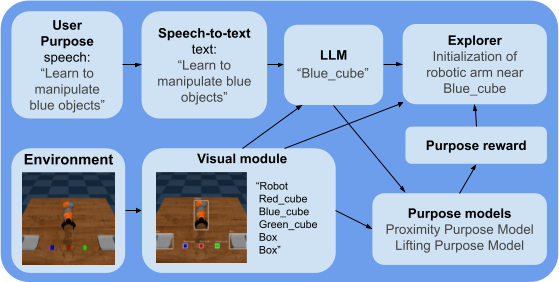}
    \end{minipage}
    \begin{minipage}{0.18\textwidth}
        \includegraphics[width=\textwidth]{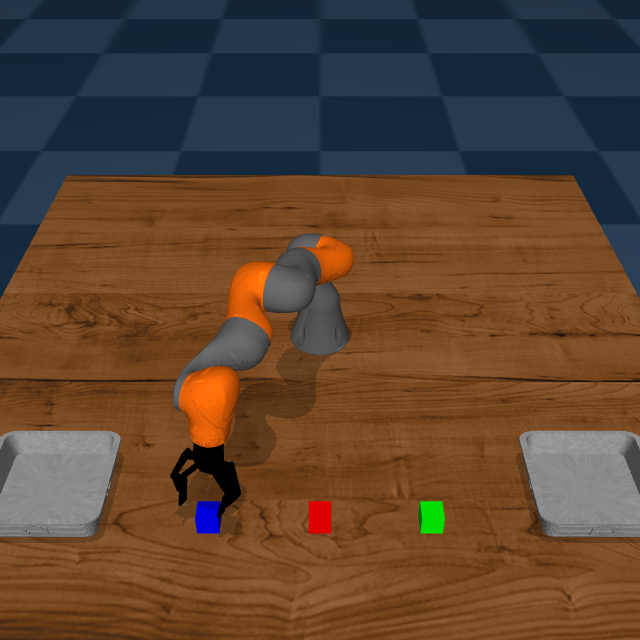}
    \end{minipage}
    \begin{minipage}{0.18\textwidth}
        \includegraphics[width=\textwidth]{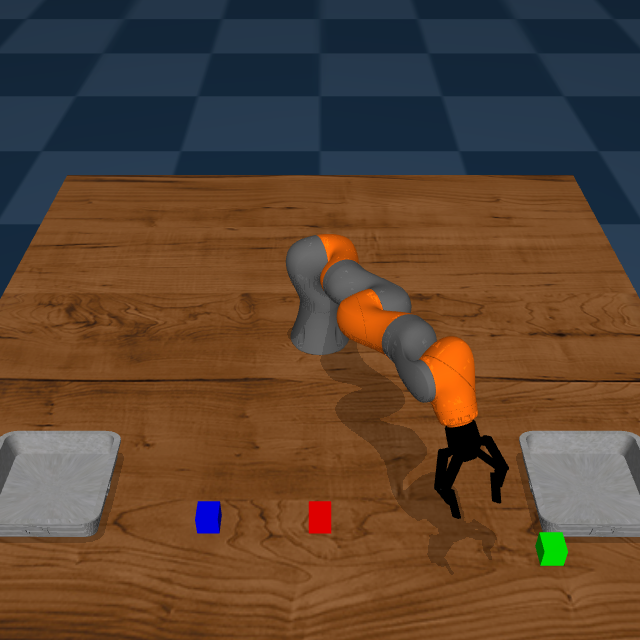}
    \end{minipage}
    \begin{minipage}{0.18\textwidth}
        \includegraphics[width=\textwidth]{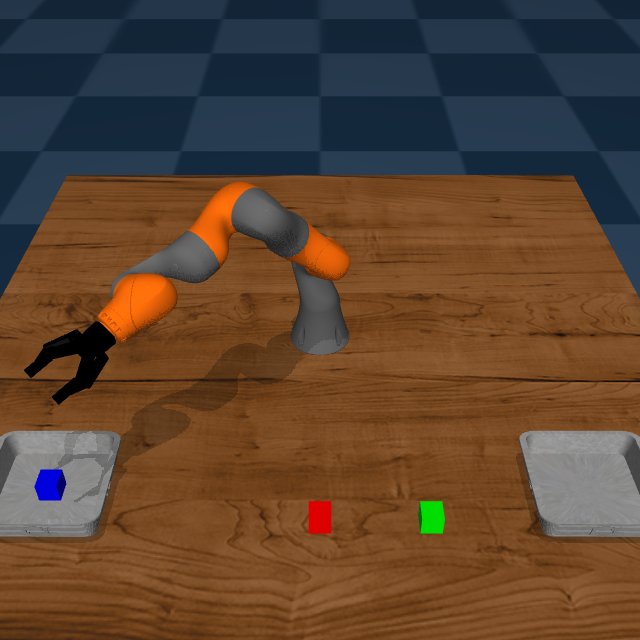}
    \end{minipage}
    \caption{Left: the POEL architecture. 
    Right: The environment comprises a robotic arm with gripper, two boxes and three objects.
    Either the blue or the green object are Purpose-related while the red object is always a distractor. 
    From left to right: 
    a "reach blue" goal, 
    a "push green" goal and 
    a "pick and place blue" goal.} 
    \label{fig:env}
\end{figure}

%

The success metrics for the different tasks vary depending on the pursued goal.
For posture tasks, the success is measured by checking if the joint positions of the robotic arm are close to the target configuration.
For pushing tasks, success is determined if the target object is within a 15 cm radius of the target position. 
Finally, for pick-and-place tasks, success is evaluated by verifying whether the target object is placed inside the correct box. 

\subsection{Performance Analysis}
The robot was evaluated with two purposes: `learn to manipulate blue objects' and `learn to manipulate green objects'. 
The results obtained were compared against both LEXA and ALAN* (the code of ALAN is not available, so we used our implementation based on LEXA plus its core aspects of visual priors and rewards based on environment changes).

\begin{figure}[h!]
    \centering
    \begin{minipage}{0.30\textwidth}
        \includegraphics[width=\textwidth]{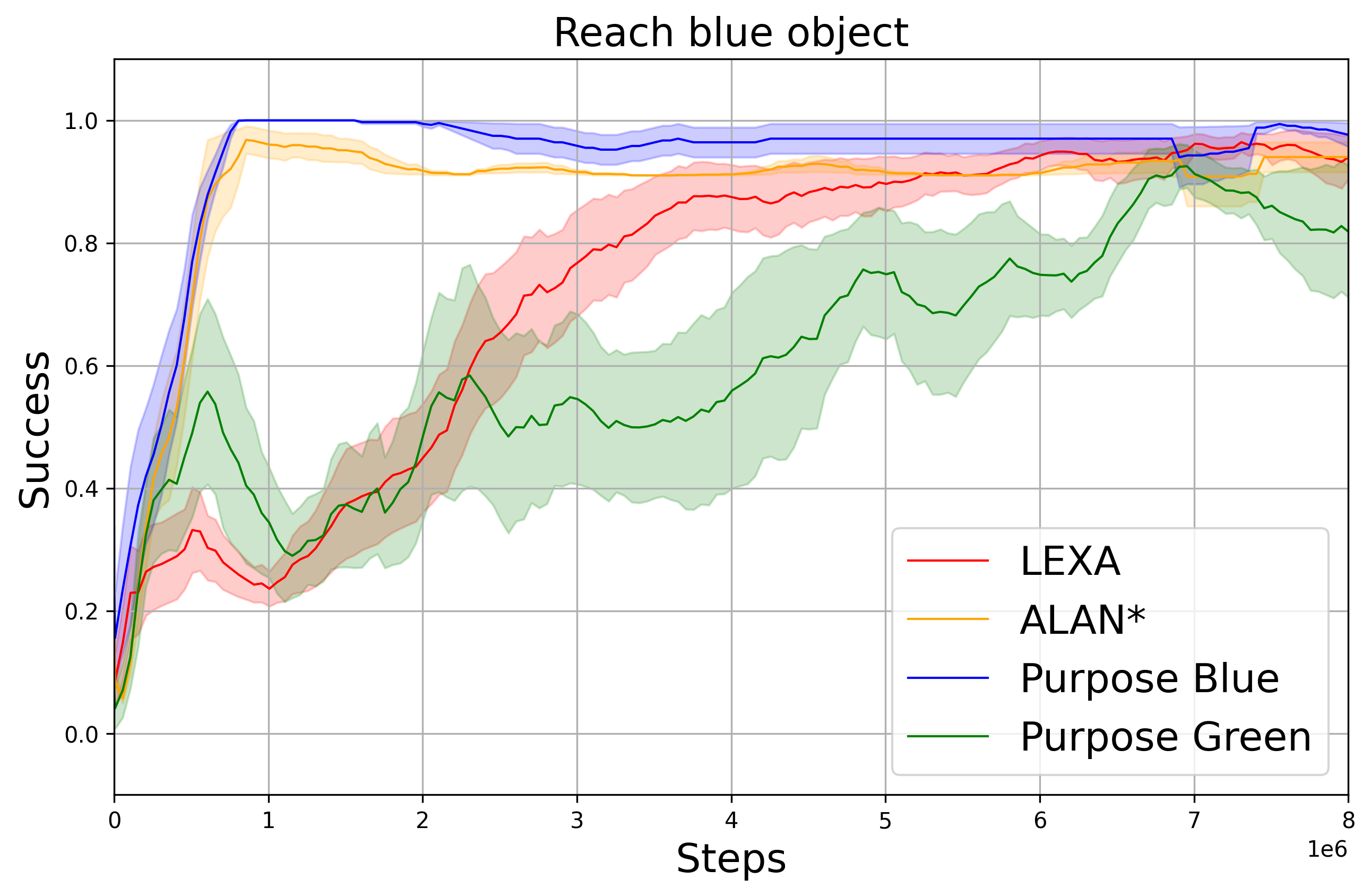}
    \end{minipage}
    \begin{minipage}{0.30\textwidth}
        \includegraphics[width=\textwidth]{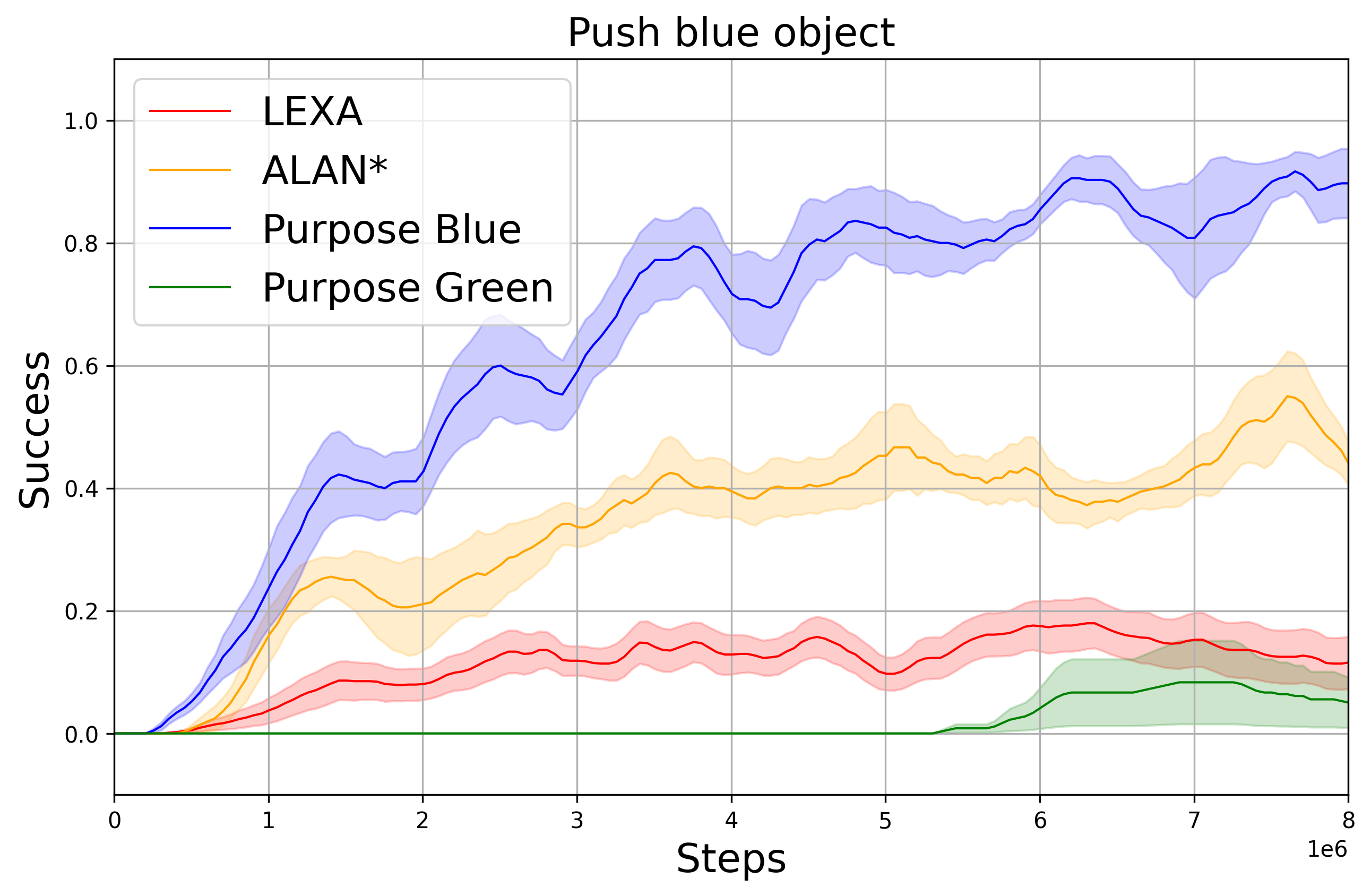}
    \end{minipage}
    \begin{minipage}{0.30\textwidth}
        \includegraphics[width=\textwidth]{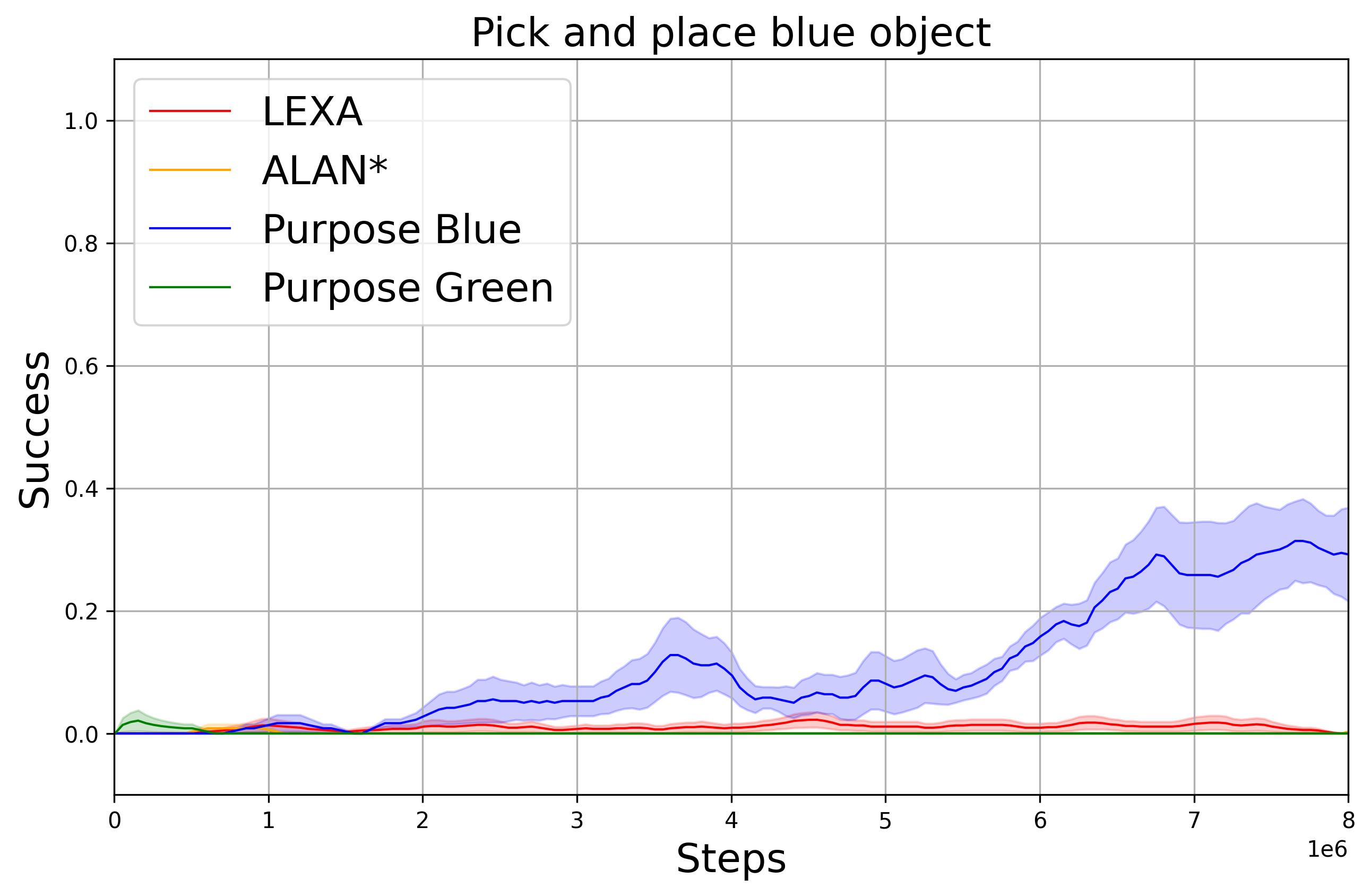}
    \end{minipage}\\
    \begin{minipage}{0.30\textwidth}
        \includegraphics[width=\textwidth]{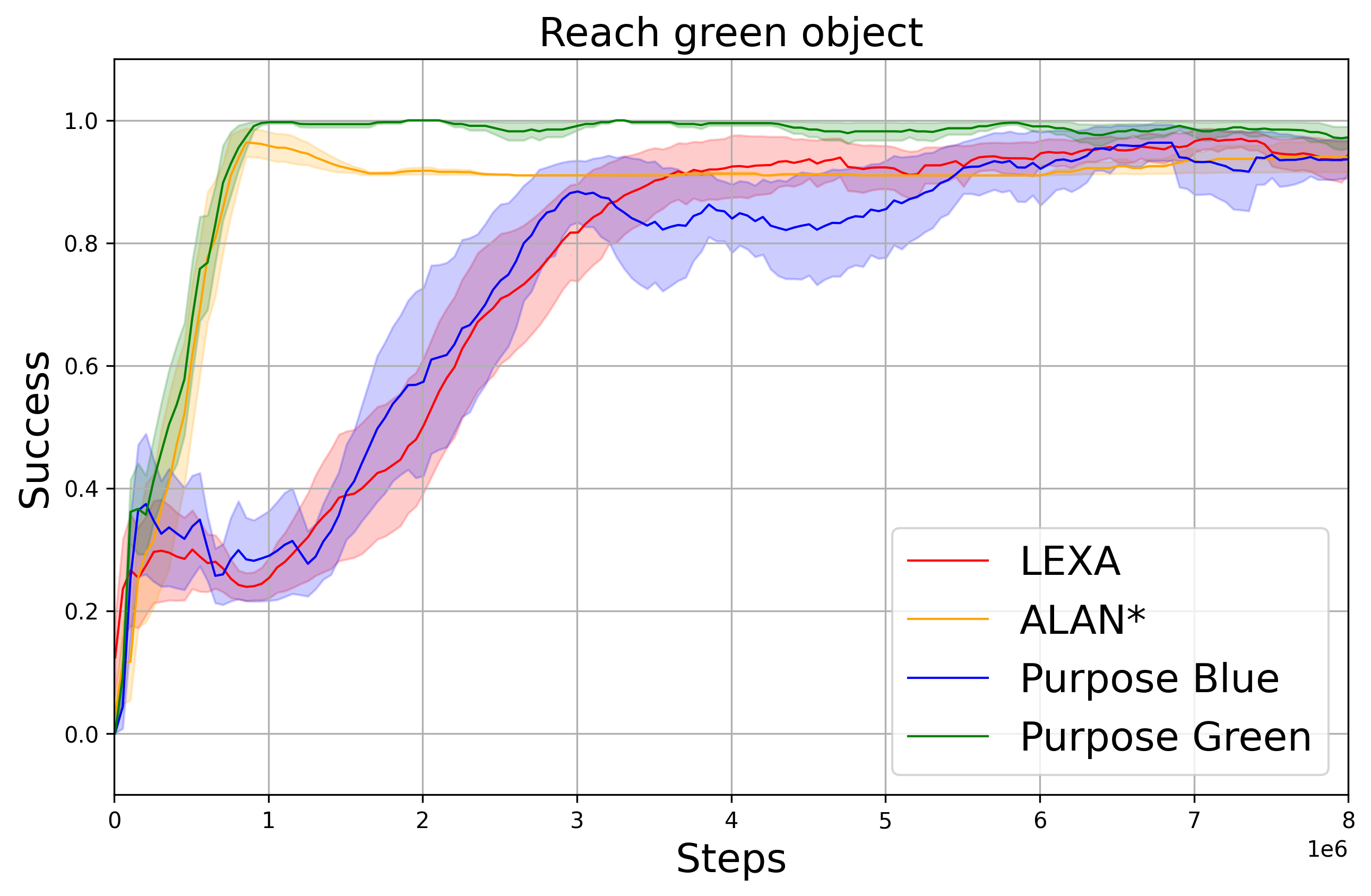}
    \end{minipage}
    \begin{minipage}{0.30\textwidth}
        \includegraphics[width=\textwidth]{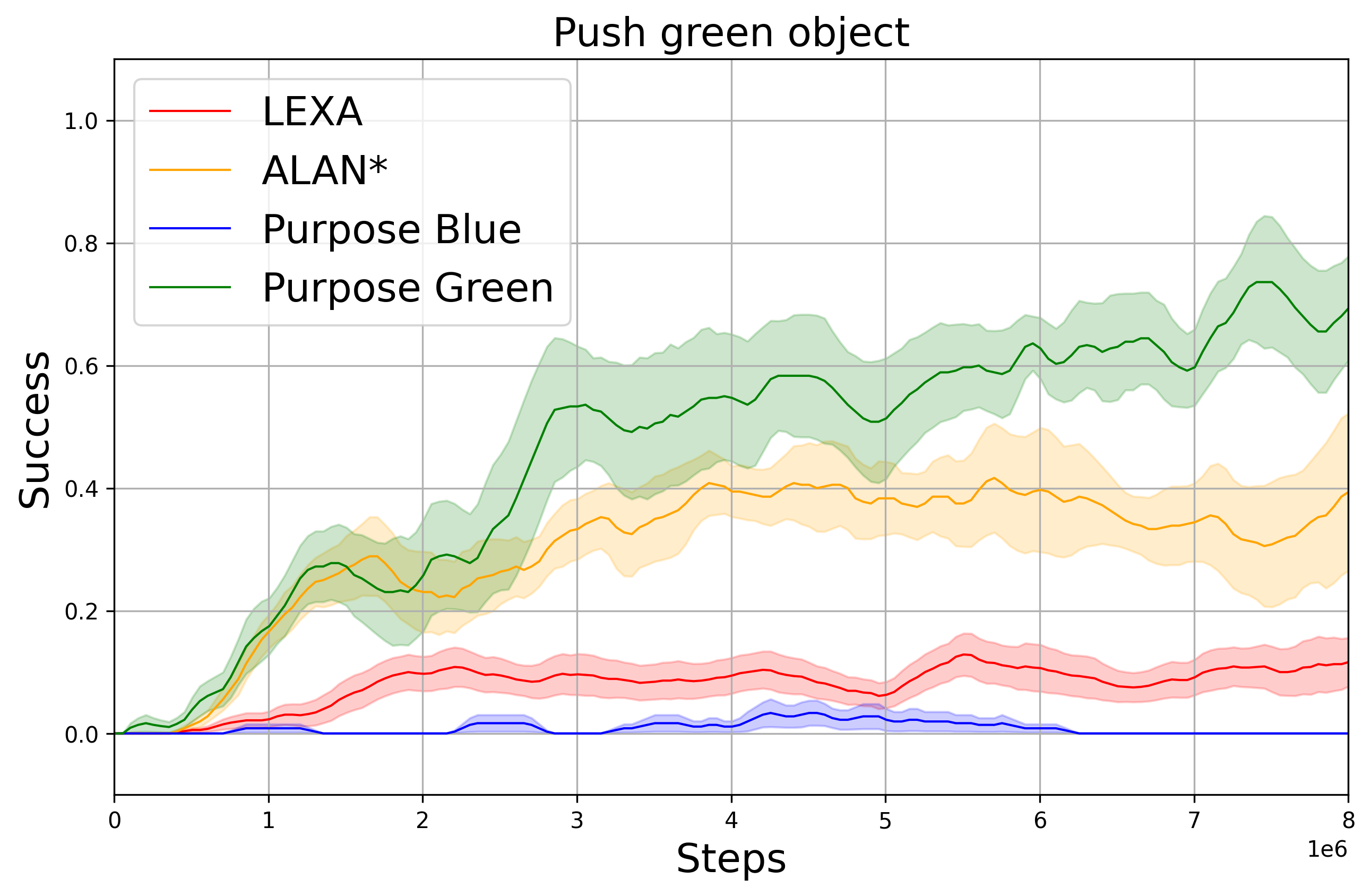}
    \end{minipage}
    \begin{minipage}{0.30\textwidth}
        \includegraphics[width=\textwidth]{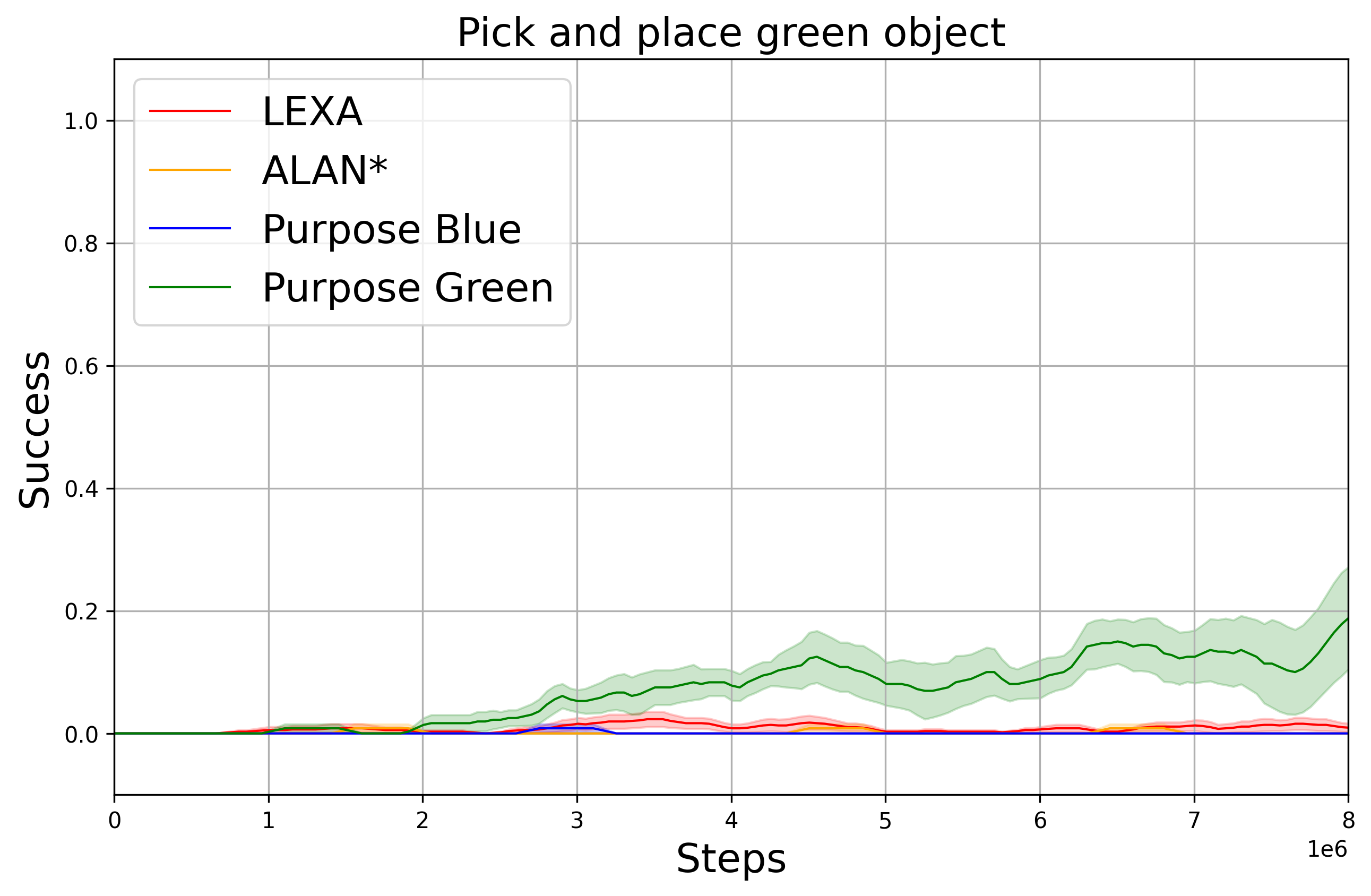}
    \end{minipage}
    \caption{Performance comparison for different types of goals: reaching, pushing, and pick and place. The pushing and pick-and-place results show averages across the 4 goals for each cube. 
    Mean performance and standard error of 3 training runs per condition, with the exception of LEXA baseline that involved 9 runs.}
    \label{fig:results_graphs}
\end{figure}

As shown in Figure~\ref{fig:results_graphs}, our algorithm significantly outperforms both LEXA and ALAN* in goals where the object being manipulated is directly correlated with the defined purpose. 
While the Purpose-directed models lose some performance on the unrelated tasks, their focus enhances the performance on Purpose-related tasks and enables them to learn tasks that LEXA or ALAN* were not able to even start to learn, such as pick and place. Note that the performance is not limited to the behaviour directly rewarded, such as staying in proximity of the object (reaching) or lifting the object, but also pushing or placing it in a box, thus showing the open-ended nature of the agent learning process. 
On the posture goals, all models perform similarly, with a slightly lower performance of ALAN* (not shown).
These results highlight the advantage of incorporating Purpose-driven exploration and targeted skill acquisition into OEL processes.

%

\section{Conclusions}
\label{Sec:Conclusions}

We introduced a novel \textit{Purpose-Directed Open-Ended Learning} (POEL) robot architecture to address a key limitation of standard OEL systems: their tendency to explore all tasks  including those irrelevant to human users. 
Leveraging the concept of \textit{purpose}, which defines what users want robots to accomplish, POEL integrates speech-to-text inputs, scene analysis, and a Large Language Model (LLM) to identify purpose-relevant objects. These objects are then used to guide exploration and self-generation of rewards, biasing learning toward tasks aligned with the user-defined purpose.

We showed the advantages of POEL in tests where a camera-arm-gripper robot interacted with both purpose-relevant and distractor objects.
The results showed that POEL significantly outperforms state-of-the-art OEL models by focusing exploration on purpose-relevant objects and achieving deeper knowledge acquisition within limited exploration time.
This marks a substantial improvement over conventional OEL approaches that dissipate resources on irrelevant tasks.

This study contributes to the state-of-the-art by operationalising the abstract concept of purpose and embedding it into OEL processes using LLM reasoning and object-centric learning biases. The proposed architecture not only enhances the applicability of OEL robots but also bridges the gap between autonomous exploration and user-specific task relevance.

Future work will focus on extending the POEL architecture to handle more complex, multi-object scenarios, refining the LLM reasoning capabilities for nuanced object-purpose associations, and evaluating the system in physical robotic platforms. Additionally, research will explore methods for dynamically adapting purposes in response to changing user needs and environmental conditions, further enhancing the versatility and autonomy of POEL robots.

\bibliographystyle{IEEEtran}
\bibliography{Bibliography}

\end{document}